\documentclass[sigconf]{acmart}

\usepackage{booktabs} 

\usepackage{adjustbox}
\usepackage{textcomp}
\usepackage{gensymb}
\usepackage{amsmath}
\usepackage{booktabs} 
\newcommand{\etal}{\emph{et al.}}
\DeclareMathOperator{\E}{\mathbb{E}}

\setcopyright{rightsretained}

\acmISBN{978-1-4503-7596-2}

\acmConference[ICVGIP'21]{12th Indian Conference on Computer Vision, Graphics and Image Processing}{December 2021}{Jodhpur, India}
\acmYear{2021}
\copyrightyear{2021}

\acmPrice{15.00}

\editor{Chetan Arora}
\editor{Parag Chaudhuri}
\editor{Subhransu Maji}

\acmDOI{10.1145/3490035.3490263}

\acmArticle{6}

\begin{document}

\title{Monocular Multi-Layer Layout Estimation for Warehouse Racks}
\titlenote{Produces the permission block, and
  copyright information}

\thanks{$^*$,$^\dagger$ denote equal contribution}
\author{Meher Shashwat Nigam}
\authornotemark[1]
\affiliation{
  \institution{Center for Visual Information Technology }
  \city{KCIS, IIIT Hyderabad}
  \country{India}
}
\email{mehershashwat@gmail.com}

\author{Avinash Prabhu}
\authornotemark[1]
\affiliation{
  \institution{Robotics Research Center }
  \city{KCIS, IIIT Hyderabad}
  \country{India}
}
\email{avinash.prabhu@students.iiit.ac.in}

\author{Anurag Sahu}
\authornotemark[1]
\affiliation{
  \institution{Robotics Research Center }
  \city{KCIS, IIIT Hyderabad}
  \country{India}
}
\email{anurag.sahu@students.iiit.ac.in}

\author{Tanvi Karandikar}
\authornotemark[2]
\affiliation{
  \institution{Robotics Research Center }
  \city{KCIS, IIIT Hyderabad}
  \country{India}
}
\email{tanvi.karandikar@students.iiit.ac.in}

\author{Puru Gupta}
\authornotemark[2]
\affiliation{
  \institution{Robotics Research Center }
  \city{KCIS, IIIT Hyderabad}
  \country{India}
}
\email{puru.gupta@students.iiit.ac.in}

\author{N. Sai Shankar}
\affiliation{
  \institution{Robotics Research Center }
  \city{KCIS, IIIT Hyderabad}
  \country{India}
}
\email{sai.nshankar@gmail.com}

\author{Ravi Kiran Sarvadevabhatla}
\affiliation{
  \institution{Center for Visual Information Technology }
  \city{KCIS, IIIT Hyderabad}
  \country{India}
}
\email{ravi.kiran@iiit.ac.in}

\author{K. Madhava Krishna}
\affiliation{
  \institution{Robotics Research Center }
  \city{KCIS, IIIT Hyderabad}
  \country{India}
}
\email{mkrishna@iiit.ac.in}

\renewcommand{\shortauthors}{}

\begin{abstract}
Given a monocular color image of a warehouse rack, we aim to predict the \textit{bird's-eye view} layout for each shelf in the rack, which we term as `multi-layer' layout prediction.
To this end, we present \textit{RackLay}, a deep neural network for real-time shelf layout estimation from a single image. Unlike previous layout estimation methods which provide a single layout for the dominant ground plane alone, \textit{RackLay} estimates the top-view \underline{and} front-view layout for each shelf in the considered rack populated with objects. \textit{RackLay}'s architecture and its variants are versatile and estimate accurate layouts for diverse scenes characterized by varying number of visible shelves in an image, large range in shelf occupancy factor and varied background clutter.  Given the extreme paucity of datasets in this space and the difficulty involved in acquiring real data from warehouses, we additionally release a flexible synthetic dataset generation pipeline \textit{WareSynth} which allows users to control the generation process and tailor the dataset according to the contingent application. The ablations across architectural variants and comparison with strong prior baselines vindicate the efficacy of \textit{RackLay} as an apt architecture for the novel problem of multi-layered layout estimation. We also show that fusing the top-view and front-view enables 3D reasoning applications such as metric free space estimation for the considered rack.
\end{abstract}

%

\ccsdesc[300]{Computing methodologies~Scene understanding}
\ccsdesc[500]{Computing methodologies~Reconstruction}
\ccsdesc[300]{Computing methodologies~Shape inference}
\ccsdesc[100]{Computing methodologies~Multi-task learning}

\keywords{Monocular 3D Reconstruction, Sim2Real, Synthetic Data Generation}

\begin{teaserfigure}
\includegraphics[width=\textwidth]{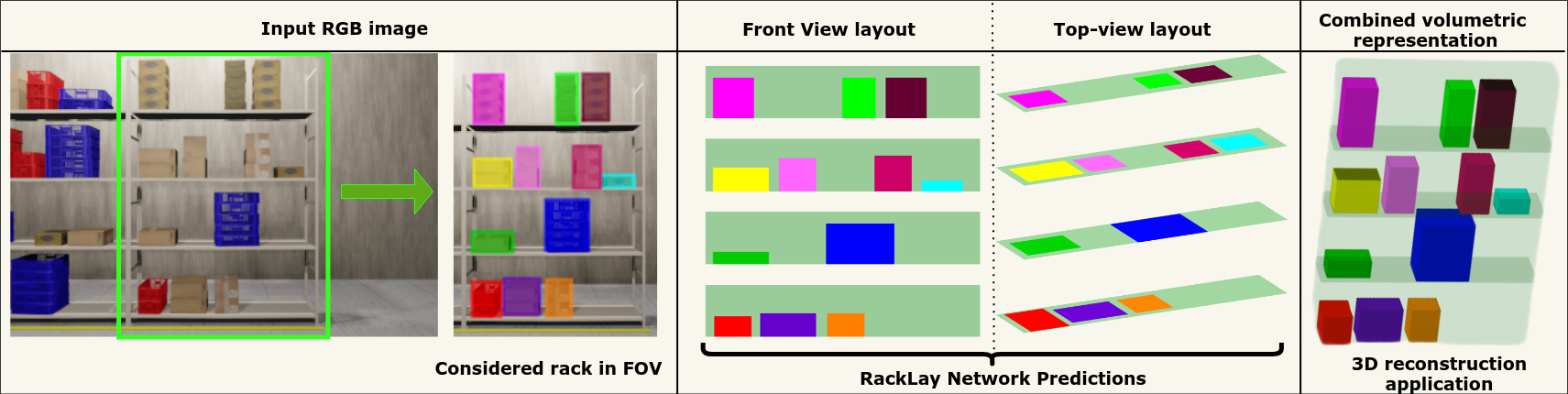}
\caption{Given a monocular RGB image of a warehouse rack, we propose \textbf{RackLay}, a deep neural  architecture that generates the \textit{top-view} and \textit{front-view} semantic layout for rack shelves and items placed on each shelf. Fusing these layouts provides a volumetric reconstruction of the rack, enabling 3D reasoning. For the considered rack in the figure, our system can report \textcolor{orange}{"Rack has 4 shelves, 12 box stacks, and 830 $cm^{3}$ of free space available"}}
\label{fig:teaser_new}
\end{teaserfigure}

\maketitle

\section{Introduction}
\label{sec:introduction}
The importance and the necessity of warehouse automation grows by the day and the future is painted with a scenario where a fleet of robots manage the warehouse with or without human intervention \cite{peter_2018}. Nonetheless, almost 30\% of the warehouses today operate without their staple warehouse management systems (WMS)\cite{wmstrends}. In such situations, essential tasks such as shelf occupancy estimation and management become important and challenging tasks.

In this paper, we address the hitherto untackled problem of layout and freespace estimation for rack shelves. This problem is equally important in the context of warehouses without WMS as well as in the scenarios where automated robotic agents 
 manipulate a shelf space. In this effort, monocular vision is the sensing modality considering the ubiquitous, low cost, high portability and scalability of such single camera systems. 
 

We propose a simple yet effective network architecture \textit{RackLay}, which takes a single RGB image as input and outputs the top-view \underline{and} front-view layouts of all the shelves comprising the dominant rack in the image (see Fig. \ref{fig:teaser_new}). \textit{RackLay} consists of a shared context encoder which reasons about the shelves and  objects together. In turn, such reasoning enables \textit{RackLay}’s decoder to generate both the \textit{top-view} and \textit{front-view} layout of the rack on a per-shelf basis (see  Fig.~\ref{fig:architecture}). 

It is important to note that the problem is not immediately reducible to any standard formulation of object recognition, layout estimation or semantic segmentation. Objects on the rack shelves are amenable for semantic segmentation~\cite{unet} or object detection~\cite{he2017mask}. However, this is not the case for racks themselves, which appear as occluded, diffused, thin structures. Indeed, these hollow structures pose a challenge for mainstream approaches. For very similar reasons, existing approaches cannot be trivially adapted for localizing rack shelves. Unlike standard layout formulations which estimate the layout with reference to a single dominant plane (e.g. ground plane)~\cite{roddick2018orthographic}, warehouse rack shelves are disconnected and distinct planar segments present at multiple heights (layers) relative to ground plane. Needless to say, these shelves can be either empty or contain an arbitrary number of objects. 
Hence, a cornerstone novelty of the present formulation is the adaptation of deep architectures to the problem of layout estimation over multiple shelf levels (layers) that constitute a rack and contents thereof.

Specifically the paper contributes as follows: 
\vspace{-1.25mm}
\begin{enumerate}
    \item It solves for the first time, the problem of shelf layout estimation for warehouse rack scenes -- a problem pertinent in the context of both warehouse inventory management as well as futuristic warehouses managed by an autonomous robotic fleet.
    \item It proposes a novel architecture, (Sec. \ref{sec:method}), the keynote of which is a shared context encoder, and most importantly a multi-channel decoder that infers the layout for each and every shelf in a given rack. We release for the first time, the \textit{RackLay} synthetic dataset consisting of 20k RGB images along with layout annotations of shelves and objects in both the top and front view.
    \item Due to the scarcity and difficulty of annotating warehouse data, we open-source a flexible synthetic data generation pipeline \textit{WareSynth} (Sec. \ref{sec:dataset_generation}) which enables the researcher/user to create and customize their own warehouse scenes and generate 2D/3D ground truth annotations needed for their task automatically. This does not restrict or limit the user to our dataset alone but provides for possibilities to create new datasets with the ability to customize as desired, as discussed in detail in Sec. \ref{sec:dataset}. 
    \item We show tangible performance gain compared to other baseline architectures \cite{shi2019pointrcnn} dovetailed and adapted to the problem of rack layout estimation. Moreover, we tabulate a number of ablations across  architectural variants which establish the efficacy and superiority of \textit{RackLay} (Sec. \ref{sec:experiments}).
    \item We also show that RackLay performs well on real-world data with the help of synthetic images generated using \textit{WareSynth} (Sec. \ref{sec:experiments:real_world}). We additionally release the \textit{RackLay} real-world dataset consisting of 542 RGB images along with layout annotations of shelves and objects in both top and front view.
\end{enumerate}
\section{Related work}
\label{sec:relatedwork}
In recent years, learning scene layouts and obtaining volumetric representations directly from an RGB image has garnered a lot of interest. Deep learning methods have become more reliable and accurate for many computer vision tasks like object detection, semantic segmentation and depth estimation. But even a combination of these fundamental solutions does not suffice for higher-order tasks like shelf-layout estimation in warehouse management systems, which requires multi-layer top-view layout estimation. To that extent, we summarize the existing approaches and differentiate our method from the rest.
\subsection{Indoor scene understanding} Room layout estimation from a single image \cite{roomnet,Lin2018IndoorSL} is a popular problem in the context of indoor 3D scene understanding. There have also been a few approaches for amodal perception as well-\cite{amodalKTCM15, factored3dTulsiani17}. Indoor scene understanding can rely on strong assumptions like a Manhattan world layout, which works well for single room layouts.
\subsection{Object detection methods}
We relate to deep learning based detection models, as a large part of our problem deals with localizing semantic classes like shelves and boxes/cartons in an 3D scene. Several existing approaches aim to detect object layouts in 3D. Some of these \cite{ku2018joint, liang2018deep} approaches combine information from images and LiDAR, others \cite{roddick2018orthographic, wang2019pseudo} work by converting images to \textit{bird’s eye view} representations, followed by object detection. 
\subsection{Bird's eye view (BEV) representation} BEV semantic segmentation has been tackled mostly for outdoor scenes. Gupta \etal \cite{Gupta2017CognitiveMA} demonstrate the suitability of a BEV representation for mapping and planning.
Schulter \etal \cite{schulter2018learning} proposed one of the first approaches to estimate an occlusion-reasoned bird’s eye view road layout from a single color image. They use monocular depth estimation and semantic segmentation to aid their network that predicts occluded road layout. 
Wang \etal \cite{wang2019parametric} build on top of \cite{schulter2018learning} to infer parameterized road layouts. Parametric models might not account for all possible scene layouts, whereas our approach is non-parametric and thus more flexible.
MonoOccupancy \cite{monooccupancy}, uses a variational autoencoder (VAE) to predict road layout from a given image, but only for the pixels present in the image. MonoLayout \cite{monolayout}, can be trained end to end on colour images, reasons beyond occlusion boundaries and does not need to be bootstrapped with these additional inputs. We predict the occlusion-reasoned occupancy layouts of multiple parallel planes(layers), with varying height, from a single view RGB image.
\subsection{Warehouse Datasets} 
\label{sec:relatedwork:dataset}
There are very few datasets publicly available for warehouse settings. Real-world datasets like LOCO\cite{loco2020} exist for scene understanding in warehouse logistics, in which they provide a limited number of RGB images, along with corresponding 2D annotations. Due to the difficulty in acquiring the 3D ground truth information from real scenes there aren't any real warehouse datasets which provide information about the objects in scene and their relative positions and orientation.
For 3D deep learning applications, large amount of diverse data along with 3D ground truth information is required. There are general purpose synthetic data simulators like NVIDIA Isaac \cite{nvidia_developer_2020}, which provide warehouse scenes.  However, they provide lesser control as to specifying properties for the warehouse, and can't be modified easily to generate annotations needed for our task. To this end, we introduce our dataset generation pipeline.
\section{Dataset Generation Pipeline}
\label{sec:dataset_generation}



In this section, we introduce our synthetic data generation pipeline termed \textit{WareSynth}, which can be used to generate 3D warehouse scenes, automate the process of data capture and generate corresponding annotations. 

\subsection{Software and initial models}

For modelling and rendering the warehouse setup, we used the open source 3D graphics toolset Blender\cite{blender}(version 2.91). We used freely available textures and 3D mesh models for creating an initial database of objects. These objects include: \textit{boxes, crates, racks, warehouse structures, forklifts, fire extinguishers} etc.
 
\subsection{Generation process}

Our generation process entails placement of objects in the scene procedurally in a randomized fashion, followed by adjustment of the lighting and textures. We perform texture editing and manipulate roughness and reflectance properties of objects to mimic real warehouse scenes.

We start with an empty warehouse. Racks are placed inside the warehouse according to a randomly generated 2D occupancy map. Lighting in the warehouse is also according to the same map, where we illuminate the corridors and also introduce ambient lighting. We keep the inter-shelf height and number of shelves in racks, width of corridors and overall rack density of the warehouse as parameters which can be tuned as per requirements. It is important to note that \textit{WareSynth} is not constrained by our specific settings. The existing models can be readily substituted with custom box and rack models to configure the warehouse. 

We also randomize the placement of boxes on each particular rack by specifying parameters which control the density of boxes placed and minimum distance between the boxes. We vary the density of rack occupancy by sampling the associated parameter from a uniform distribution between 0 (empty shelf) and 1 (fully occupied shelf). This ensures that the data is not imbalanced. Our algorithm picks a random box from available box models, and positions the same at a random angle varying between $\pm r^{\circ}$ about the vertical axis (where $r$ can be specified). The boxes can also be stacked over each other, on individual shelves. 
This probabilistic procedure ensures that boxes are placed on the shelves randomly, but within the specified constraints, which helps us generate a large variety of realistic data.

\subsection{Data capture and annotation}

We capture data via movement of a 6-DoF camera around the warehouse corridors by specifying a path or a set of discrete positions. The camera parameters can be varied in order to produce a diversity of views. The camera rotation, focal length, height above the ground etc. can all be manipulated and constrained according to the kind of views desired.

As per the requirement, we can capture the RGB images at the desired resolution for each of these camera positions, along with the camera intrinsic, and extrinsic parameters. We can also extract 2D annotations such as 2D bounding boxes and semantic and instance segmentation masks of the objects. By using this pipeline on a NVIDIA RTX 2080Ti we are able to generate 80 images per-minute. We can also obtain the 3D positions, orientations and 3D bounding boxes for all objects present in the camera FOV, along with depth maps and normal information. Our pipeline can also be used to obtain stereo-information. The obtained data can be easily be exported to various popular annotation formats such as KITTI, COCO, Pix3D, BOP etc.

\subsection{Applications and extensions}

\textit{WareSynth} can be used for various tasks such as 2D/3D object detection, semantic/instance segmentation, layout estimation, 3D scene navigation and mapping, 3D reconstruction, etc. The same pipeline can also be modified to other kinds of scenes such as supermarkets, greenhouses, etc., by changing the database of objects and placement parameters. The generation procedure and data capture methods are efficient, very flexible and can be customized as per user requirement. This makes the pipeline very useful for future research and generating annotated data at a large scale.
\begin{figure*}[!ht]
    \center
    \includegraphics[height=7cm]{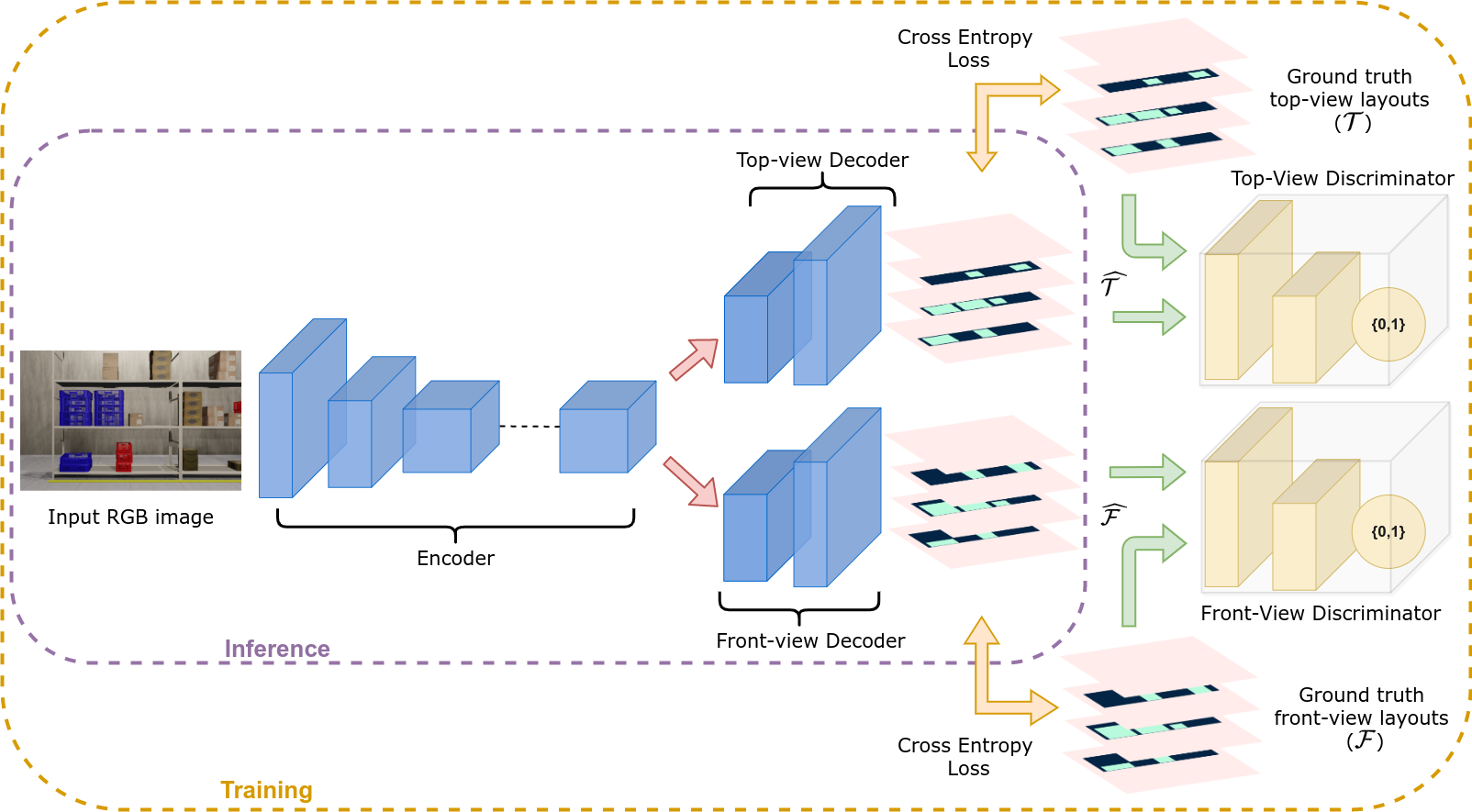}
    \caption{\small{The figure shows architecture diagram for \textit{RackLay-D-disc}. It comprises of a context encoder, multi-channel decoders and adversarial discriminators. (Refer to Sec. \ref{sec:experiments:eval}).}}
    \label{fig:architecture}
\end{figure*}

\section{Method}
\label{sec:method}

\subsection{Problem Formulation}
\label{sec:method:problemform}

Formally, given a monocular color image $\mathcal{I}$ of a warehouse rack in perspective view, we aim to predict the top-view (bird's eye view) layout for each shelf of the rack that lies within distance \textbf{\textit{d}} from the camera (range of detection) and is visible in the image\footnote{Flat-earth assumption: For the scope of this paper, we assume that the concerned rack is located within a bounded area in front of the camera, and that all planes(layers) in consideration (rack shelves, ground etc.) are more or less planar, with almost no inclination. The rack shelves are considered to lie on horizontal planes(layers) parallel to the ground plane.}. 

Concretely, we wish to learn a labelling function that generates a top-view layout for the set of all scene points within a region of interest $\Omega$. Here, we consider $\Omega$ to be a rectangular area around the concerned rack's center, in a top-down orthographic view of each shelf plane. The labeling function must produce labels for all the points in $\Omega$, regardless of whether or not they are imaged in $\mathcal{I}$. 
The points in $\Omega$ are labelled as \textit{occupied}, \textit{unoccupied} or \textit{background}. In our problem context, a pixel with \textit{occupied} label denotes the presence of an object (boxes, cartons, crates etc.) at that place on the shelf, and the \textit{unoccupied} label denotes that the shelf is empty at that pixel. Label \textit{background} denotes the area that is not occupied by the shelf. As an additional task, we aim to learn a similar labelling function for the points in the front-view layout, which is orthogonal to the top-view layout. Here, we classify the empty inter-shelf area as \textit{unoccupied}. Using a combined representation from these layouts, we obtain a 3D reconstruction of the rack, which can be further used for 3D spatial reasoning tasks. We discuss this extension in the further sections. 

\subsection{RackLay Architecture}

The architecture of RackLay comprises of the following subnetworks (refer Fig. \ref{fig:architecture}):
\begin{enumerate}
    \item A \textbf{context encoder} which extracts relevant 3D scene and semantic features from the input RGB image $\mathcal{I}$ for layout estimation. This provides a context $\mathcal{I}_e$ that helps us identify \textit{occupied}, \textit{unoccupied} and \textit{background} scene points, for each shelf in the rack, in subsequent processing. We use a ResNet-18 encoder (pre-trained on ImageNet\cite{deng2009imagenet}) and fine-tune this feature extractor to learn low-level features that help reason across the three scene classes.
    
    \item A \textbf{top-view decoder} that can comprehend the context to generate layouts for each shelf of the rack that is visible in the image. It decodes the context from the feature extractor (context encoder) via a series of deconvolution and upsampling layers that map the context to a semantically rich bird’s eye view. The decoder outputs an $\mathcal{R}$ × $\mathcal{D}$ × $\mathcal{D}$ grid which represents the top-view layout $\mathcal{T}$ where $\mathcal{R}$ is the number of output channels and $\mathcal{D}$ × $\mathcal{D}$ is the resolution for the output layouts. Each channel represents a shelf and is a per-pixel label map of the corresponding top-view layout. It is important to note the novelty of the associated design choice, i.e. using a multi-channel output to predict occupancy layouts which lie at different heights (layers) in the rack.

    \item A \textbf{discriminator} is an adversarial regularizer.
    It refines the predicted layouts by regularizing their distributions to be similar to the true distribution of plausible layouts. The layouts estimated by the top-view decoder are input to this patch-based discriminator. The discriminator regularizes the distribution of the output layouts (\textit{fake} data distribution, in GAN\cite{GAN} parlance) to match a prior data distribution of conceivable scene layouts (\textit{true} data distribution). 
\end{enumerate}

In order to deduce both views (top, front) from a single unified model, we extend the above architecture by adding an identical decoder to the existing encoder, followed by a discriminator, which predicts front-view layout for each shelf ($\mathcal{F}$), just like the top-view layout ($\mathcal{T}$). 

\subsection{Formulation}

 We formulate the multi-rack shelf layout estimation as a multi-task probabilistic inference problem. Formally, let the top-view layout tensor be ${\mathcal{T}}$ and the front-view counterpart be ${\mathcal{F}}$. We configure a deep neural network to maximize the posterior distribution $P({\mathcal{T}},{\mathcal{F}} | \mathcal{I})$ given an RGB image $\mathcal{I}$. $\mathcal{R}$ is a flexible parameter and denotes the maximum number of shelves the network can detect. Let ${\mathcal{T}}_{i}$ and ${\mathcal{F}}_{i}$ represent $\mathcal{D}$ × $\mathcal{D}$ per-pixel occupancy label maps for the $i^{th}$ shelf of the rack ($i\in 1,2,\ldots \mathcal{R}$). ${\mathcal{T}}_{i}$ and ${\mathcal{F}}_{i}$ are shelf-centric as well as aligned with the shelf’s coordinate frame. 
Conditioned on the encoded features $\mathcal{I}_{e}$ of the input image $\mathcal{I}$, marginals $\mathcal{T}$ and  $\mathcal{F}$ are independent of each other. Additionally, the components ${\mathcal{T}_i}$ and  ${\mathcal{F}_i}$ are all independent of each other since the occupancy of each  shelf does not depend on other shelves in the rack. Consequently, for the combined task, the posterior can be factorized as follows:
\vspace{-1mm}
\begin{align*}
    P({\mathcal{T}}, {\mathcal{F}} &\lvert \mathcal{I}) \\ 
    &= \begin{aligned} P({\mathcal{T}} \lvert \mathcal{I}_{e}) P({\mathcal{F}} \lvert \mathcal{I}_{e}) \end{aligned}\\
    &= \begin{aligned}P({\mathcal{T}}_{1}, {\mathcal{T}}_{2} \ldots {\mathcal{T}}_{\mathcal{R}} \lvert \mathcal{I}_{e}) P({\mathcal{F}}_{1}, {\mathcal{F}}_{2}...{\mathcal{F}}_{\mathcal{R}}  \lvert \mathcal{I}_{e}) \end{aligned} \\
    &= \begin{aligned}\underbrace{ \prod_{i=1}^{\mathcal{R}} P({\mathcal{T}_i} \lvert \mathcal{I}_{e})}_\text{top-view decoder} \hspace{2mm}       
     \underbrace{\prod_{i=1}^{\mathcal{R}} P({\mathcal{F}_i} \lvert \mathcal{I}_{e})}_\text{front-view decoder} \end{aligned}
\end{align*}

\subsection{Loss function}

The network parameters $\phi$, $\psi$, $\theta$ of the context encoder, the top-view decoder and discriminator respectively are optimized using stochastic gradient descent.

\[\mathcal{L}_{sup}(\widehat{\mathcal{T}};\phi, \psi) = \sum_{j=1}^{N}\sum_{i=1}^{\mathcal{R}} f\left( \widehat{\mathcal{T}}_{i}^{j}, \mathcal{T}_{i}^{j} \right)\]
\[\mathcal{L}_{adv}(\widehat{\mathcal{T}};\phi, \psi, \theta) = \E_{\theta\sim p_{fake}}[(\widehat{\mathcal{T}}(\theta)-1)^{2}] \]
\begin{align*}
 \mathcal{L}_{discr}(\widehat{\mathcal{T}};\theta) = \E_{\theta\sim p_{true}}[(\widehat{\mathcal{T}}(\theta)-1)^{2}] \\ +
\E_{\theta\sim p_{fake}}[ (\widehat{\mathcal{T}}(\theta)-0)^{2}] 
\end{align*}

where, $\widehat{\mathcal{T}}$ and $\mathcal{T}$ are the predicted and the ground truth top-view layouts for each shelf, $\mathcal{R}$ is the maximum number of shelves considered and $N$ is the mini-batch size. 

\textbf{$\mathcal{L}_{sup}$} is the standard per-pixel cross entropy loss which penalizes deviation of the predicted layout labels ($\widehat{\mathcal{T}}$) from their corresponding ground-truth values ($\mathcal{T}$). The adversarial loss $\mathcal{L}_{adv}$ encourages the distribution of layout estimates from the top-view decoder ($p_{fake}$) to be close to the true data distribution ($p_{true}$).  $\mathcal{L}_{discr}$ enforces the discriminator to accurately classify the network generated top-view layouts from the layouts sampled from the true data distribution. The discriminator loss $\mathcal{L}_{discr}$ is the discriminator update objective\cite{GAN}. Note that a similar set of loss terms exist for front-view layout estimation as well.
\section{Experiments and Analysis}
\label{sec:experiments}

\begin{table*}[!ht]
\begin{center}
\begin{adjustbox}{max width=\textwidth}
\begin{tabular}{c|c|c|c|c|c|c|c|c}
& \multicolumn{4}{c|}{\textbf{Top View}} 
& \multicolumn{4}{c}{\textbf{Front View}} \\ 
\cline{2-9}
& \multicolumn{2}{c|}{\textbf{Rack}} 
&   \multicolumn{2}{c|}{\textbf{Box}}    
& \multicolumn{2}{c|}{\textbf{Rack}}   
& \multicolumn{2}{c}{\textbf{Box}}  \\ 
\hline
\textbf{Method}       & \textbf{mIoU}      &     \textbf{mAP}           &  \textbf{mIoU}       & \textbf{mAP}         &   \textbf{mIoU}    &   \textbf{mAP}   &      \textbf{mIoU}   &   \textbf{mAP}   \\
\cline{1-9} 

RackLay-D-disc  & $93.15$   &      $\textbf{98.73}$      & $\textbf{95.07}$     &  $97.90$      &   $90.75$  &  $\textbf{98.54}$   &   $94.29$  &  $\textbf{97.95}$     \\  
RackLay-D  & $\textbf{95.03}$   &      $98.37$      & $92.94$     &  $97.63$    &  $\textbf{95.21}$  &  $98.48$   &   $\textbf{95.17}$   &  $97.94$         \\
RackLay-S-disc  & $92.34$   &      $98.28$      & $93.71$     &  $97.85$    &   $91.96$  &   $98.13$   &   $92.65$   &  $97.51$     \\
RackLay-S  & $93.02$   &      $98.61$      & $94.61$     &  $\textbf{98.07}$    &   $94.30$  &   $98.09$   &  $92.11$ &  $97.56$         \\
                              
\hline
PseudoLidar-PointRCNN\cite{shi2019pointrcnn}   & $73.28$   &      $77.40$      &  $55.77$     &  $81.26$    &   $-$   & $-$   &     $63.05$   &  $89.45$     \\

MaskRCNN-GTdepth\cite{he2017mask}   & $36.48  $   &      $42.48$      &  $35.57 $     &  $47.44$    &   $-$   & $-$   &     $-$   &  $-$     \\
     
\end{tabular}
\end{adjustbox} 
\end{center}
\caption{\small{Quantitative results: We benchmark the 4 different versions of our network- \textit{RackLay-S}, \textit{RackLay-S-disc}, \textit{RackLay-D} and \textit{RackLay-D-disc}, along with two baselines- \textit{PseudoLidar-PointRCNN}\cite{wang2019pseudo,shi2019pointrcnn} and \textit{MaskRCNN-GTdepth}\cite{he2017mask} (as described in Sec. \ref{sec:experiments:eval}). Note that \textit{RackLay-S} and \textit{RackLay-S-disc} are single decoder models and hence cannot predict top view and front view simultaneously. The top view and front view results displayed for each of these two models were trained separately (scaled results out of 100).}}
\label{table:quantitative:main}
\end{table*}

\subsection{RackLay Dataset}
\label{sec:dataset}

For the purpose of training and testing our network, using \textit{WareSynth}, we generated 2 datasets \footnote{Download RackLay datasets:\url{http://bit.ly/racklay-dataset}}, a simple dataset with 8k images and a complex dataset with 12k images. We describe and display results for our more diverse and complex dataset consisting of 12k images, which we split into 8k/2k/2k for train/test/validation. We introduced two kinds of variations during data generation to add diversity and complexity in the scenes such that the resulting scenes mimic counterparts from real warehouses (see Fig. \ref{fig:diversity}). We describe these variations below.

\noindent \textbf{Scene-level variation:} Each object placed on the racks is randomly chosen from a set of $6$ distinct cardboard box categories and $2$ different types of colored crates. These items all have different dimensions, textures and reflective properties (observe rows 1, 2 and 4 of Fig. \ref{fig:diversity}). We also vary the inter-shelf height and the rack width between different scenes. The height up to which objects can be stacked over each other is also randomized (observe rows 2, 4 and 6 of Fig. \ref{fig:diversity}). The background for the racks can be a wall (observe rows 1, 2 and 4 of Fig. \ref{fig:diversity}) or other locations of a busy warehouse (observe rows 3, 5, 6, and 7 of Fig. \ref{fig:diversity}), possibly containing other racks. Note that this setting poses a challenge for layout estimation since the network now needs to differentiate the concerned rack of interest from racks and other distractions in the background.

\noindent \textbf{Variation in camera placement:} The camera is placed such that it is directly facing the racks, and the image plane is orthogonal to the ground plane. For the camera, its horizontal distance to the rack and its vertical position above the ground plane are two parameters that are varied. Applying these variations affects the number of shelves visible in the image from 1 to $\mathcal{R}$. For our dataset, we set $\mathcal{R}$=4 (observe rows 1, 2, 3 and 4 of Fig. \ref{fig:diversity}).

\noindent \textbf{Ground-truth layout generation:} For every considered camera position, we record the corresponding ground truth information (location, dimensions, object type) for all objects appearing within the intersection of the camera field of view (FOV) and considered range of detection as defined in the problem formulation (\ref{sec:method:problemform}). 

From this information, we generate ground truth top-view and front-view layouts by projecting onto the horizontal and vertical planes in the shelf-centric reference frame as described earlier. In our setting, the top-view and front-view layouts are $512 \times 512$ 2D pixel  grids (hence $\mathcal{D}=512$) which map to a corresponding $8m \times 8m$ spatial extent. Therefore, the spatial resolution comes out be 1.5625cm/pixel. This mapping helps us estimate free space in metric 3D. 

\subsection{Evaluated Methods and Metrics}
\label{sec:experiments:eval}
We evaluate the performance for the following approaches:
\begin{itemize}
    \item \textit{PseudoLiDAR-PointRCNN}: A PointRCNN based architecture \cite{shi2019pointrcnn}, for 3D object detection on PseudoLiDAR \cite{wang2019pseudo} input, which involves converting image-based depth maps to LiDAR format. The 3D object detections are projected to the horizontal plane to obtain BEV layouts. We chose PointRCNN due to the success it enjoys in bird's eye view 3D object detection tasks.
    \item \textit{MaskRCNN-GTdepth}: Instance segmentation method \cite{he2017mask}, paired with ground-truth depth maps.  
\end{itemize}
We compare that with following variants of RackLay:
\begin{itemize}
    \item \textit{RackLay-S}: Single decoder architecture, can be either for front-view \textit{or} top-view.
    \item \textit{RackLay-D} : Double decoder architecture, for both front-view \textit{and} top-view.
    \item \textit{RackLay-S-disc} : Single decoder architecture with discriminator, can be either for front-view \textit{or} top-view.
    \item \textit{RackLay-D-disc} : Double decoder architecture with discriminators, for both front-view \textit{and} top-view.
\end{itemize}
 We evaluate the layouts on both Mean Intersection-Over-Union (mIoU) and Mean Average-Precision (mAP). These metrics are calculated on a per-class basis.
\subsection{Results}

We started with a standard encoder-decoder \textit{RackLay-S} architecture, for predicting the top-view layouts. Having achieved superior results as compared to the baselines, we trained an identical architecture for predicting front-view layouts, which also gave similar results as top-view (refer Table \ref{table:quantitative:main}). To obtain both top and front views simultaneously in a single forward pass, we trained a double decoder model \textit{Racklay-D} for estimating both top-view and front-view. Here, we observed gains in performance both quantitatively (refer row 1 in Table \ref{table:quantitative:main}) and qualitatively (refer to Fig. \ref{fig:singlevsdouble}). We further improved upon \textit{Racklay-D} with adverserial regularization to get our best network, \textit{Racklay-D-disc}, which gives cleaner and sharper layouts, as discussed in ablation studies (refer Sec. \ref{sec:experiments:ablation}).

In Fig. \ref{fig:diversity}, observe how our best network \textit{RackLay-D-disc} is able to estimate layouts for a variety of scenes. For varying number of shelves (rows 1-4), we are able to predict layouts that are visible in the image and an empty layout for the rest. Our dataset also contains extremely sparse and densely packed shelves (rows 4-5), for which our network is able to reason out the thin spaces. We are also able to reason between the concerned rack and background clutter. Images with background clutter are shown in rows 3, 5, 6 and 7 and images with a wall behind are shown in rows 1, 2 and 4. Our network is also able to estimate layouts with for the case where no boxes are placed on a shelf as shown in row 7.
\subsection{Comparison with baselines}

\subsubsection{\textit{PseudoLiDAR-PointRCNN}}
We perform 3D object detection using PointRCNN\cite{shi2019pointrcnn} on a PseudoLiDAR input\cite{wang2019pseudo}. The PointRCNN architecture was designed for detecting objects on a road like scene, their approach assumes a single dominant layer (the ground plane). This is an assumption used by many methods designed for bird's eye view, and hence may not perform well for indoor scenes where multiple objects are scattered at different heights relative to ground plane. We observed that the success enjoyed by PointRCNN does not translate in the presence of multi-layer data. Their network is able to identify only the bottom shelf and objects kept on it. Therefore, we report metrics only for the bottom shelf layouts (refer Table \ref{table:quantitative:main}). This again highlights the importance of our work because we reason about bird's eye view representation for multiple layers, rather than a single dominant layer. 
\subsubsection{\textit{MaskRCNN}} We also compare with a classical approach wherein we perform instance segmentation using MaskRCNN\cite{he2017mask}, and pair it with ground truth depth-maps. Leveraging the depth maps, we project the detected boxes to 3D, shelf-wise, using the fact that boxes on a particular shelf will have similar vertical image coordinate. We then re-project the obtained points on to the horizontal plane to obtain the top-view box layouts for each shelf, by computing a convex hull for each box. Since this approach can only reason about visible points in the image, it is clear from Table \ref{table:quantitative:main} that our network performs much better as it is able to perform amodal perpection and is able to complete shapes and parts of the layout unseen in the input image. As discussed in Sec. \ref{sec:introduction}, MaskRCNN fails to predict segmentation maps for thin structures like shelves with good accuracy. Therefore, we present results only for box layouts.
\subsection{Ablation studies}
\label{sec:experiments:ablation}
We conduct ablation studies to analyze the performance of various components in the pipeline. These lead to a number of interesting observations.
\begin{figure}[!t]
\centering
\includegraphics[width=0.8\columnwidth]{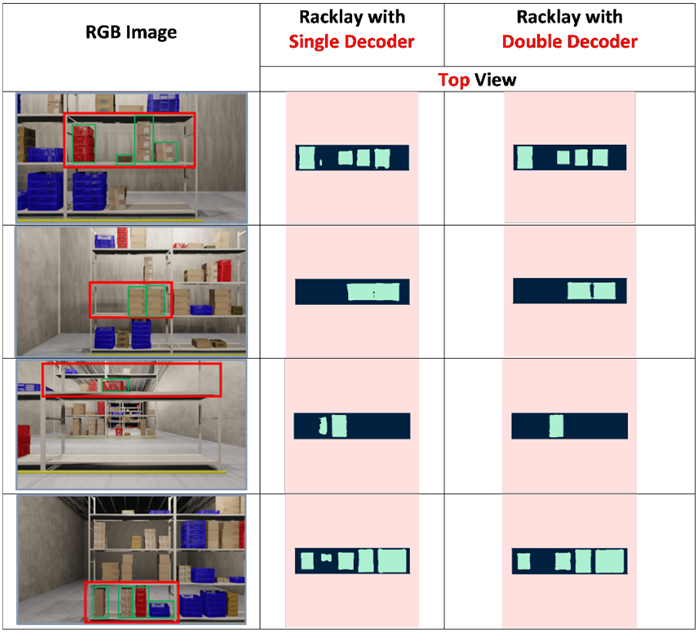}
\caption{\small{\textbf{Effect of dual-task learning on top-view layout estimation}: The output layouts are being displayed only for the shelf bounded with a red box. The corresponding boxes for the bounded shelf are bounded with green boxes. Observe how the predicted boxes in \textbf{column 3} are more box-like and the presence of spurious noise in box predictions is lesser \textbf{(row 1 and 4)}. The double-decoder model performs better in case where there are boxes very close to each other \textbf{(row 2)} and does a better job predicting the space between the boxes. It is also better at differentiating between boxes in the background and foreground \textbf{(row 3)}. The single-decoder model confuses the background box to the left of the actual foreground box to be in the foreground as well, however the double-decoder model avoids this. The double-decoder model also avoids predicting spurious boxes as shown in \textbf{(row 4)}.}}
\label{fig:singlevsdouble}
\end{figure}

\subsubsection{\textbf{Multi-task learning}}
For the additional task of estimating the front-view layouts of the shelves, we added a decoder to the existing encoder, which hence becomes the shared representation. Upon training our model for the dual task of front-view and top-view layout estimation, we observed that the losses were converging faster as opposed to the single decoder model, with a slight quantitative improvement (refer Table \ref{table:quantitative:main}) in performance for almost all cases. There is a considerable improvement qualitatively as shown in Fig. \ref{fig:singlevsdouble}, the network avoids predicting spurious boxes and outputs cleaner layouts. Making the network learn these two tasks together forces it to learn the relevant features related to the occupancy much faster and improves the performance metrics. 

\begin{figure}[!t]
\centering
\includegraphics[width=0.8\columnwidth]{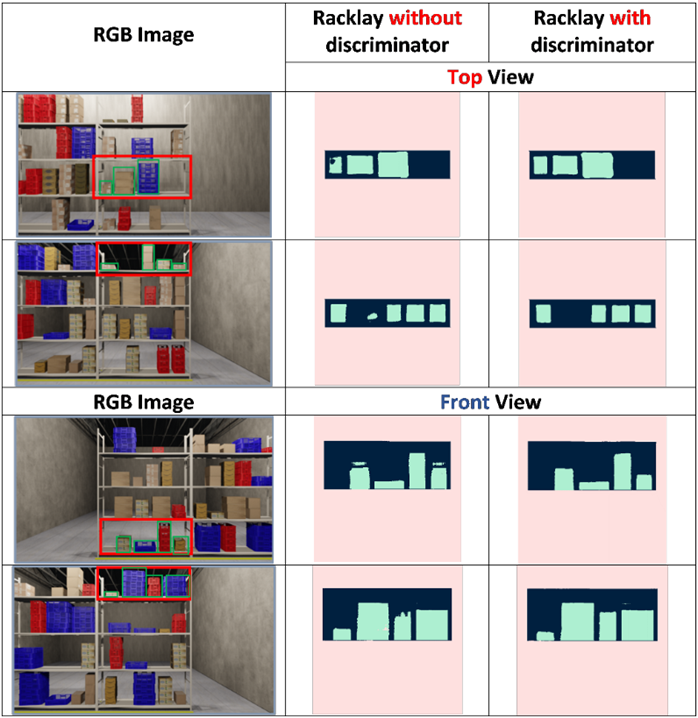}
\caption{\small{\textbf{Effect of Discriminator on qualitative performance}: The output layouts in \textbf{(columns 2-3)} are being displayed only for the shelf bounded with a red box in \textbf{(column 1)}. The corresponding boxes for the bounded shelf are bounded with green boxes. Observe how using a discriminator leads to more filled out layouts for boxes. The model without the discriminator tends to predict boxes with holes in them \textbf{(rows 1, 3 and 4)}, whereas the discriminator ensures that this does not happen. The model with the discriminator also avoids false predictions for boxes as seen in \textbf{row 2}. }}
\label{fig:discvsnodisc}
\end{figure}
\subsubsection{\textbf{Adversarial learning}}
We add a discriminator at the end of the decoder to improve the layouts.
At first glance, from Table \ref{table:quantitative:main}, using discriminators does not seem to produce any significant improvements quantitatively. However, there is considerable improvement qualitatively as shown in Fig \ref{fig:discvsnodisc}, we obtain much sharper and realistic layouts. Most notably in the case of estimating layout for boxes, use of a discriminator reduces stray pixels which are mis-classified as boxes and outputs more clean box-like structures. Adding this component enhances the capability of the network to capture the distribution of \textit{plausible} layouts.

\begin{figure*}
    \centering
    \includegraphics[width=0.88\textwidth]{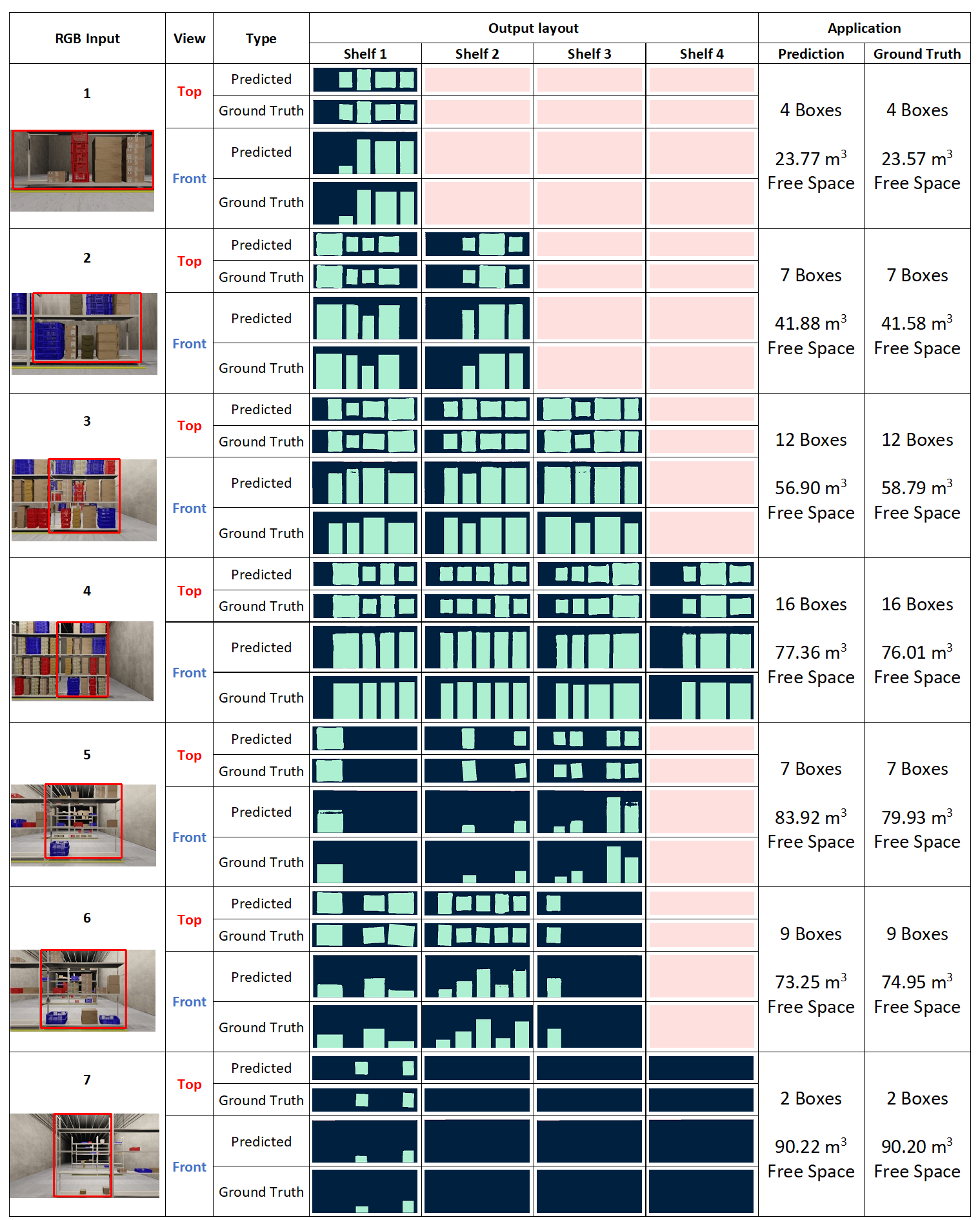}  \label{fig:diversity}
    \caption{Diversity in dataset:
    The output layouts are being displayed only for the rack bounded with a red box in \textbf{column 1}. Here, the objects on the shelves in both top-view and front-view are in green color, the free space is in dark blue color. The background is represented with pink. For the sake of visualisation, we have omitted the background class for visible shelves.  Observe how our dataset contains images of varying number of shelves and boxes\textbf{(rows 1-4)}, extremely sparse and densely packed shelves \textbf{(rows 4-5)} and scenes with background \textbf{(rows 3, 5, 6, and 7)} and without \textbf{(rows 1, 2 and 4)} background clutter.}
    \label{fig:diversity}
\end{figure*}

\subsection{Results on Real-World Images} 
\label{sec:experiments:real_world}

\begin{table}[!ht]
\begin{center}
\begin{adjustbox}{max width=\columnwidth}
\begin{tabular}{c|c|c|c|c|c|c|c|c}
& \multicolumn{4}{c|}{\textbf{Top View}} 
& \multicolumn{4}{c}{\textbf{Front View}} \\ 
\cline{2-9}
& \multicolumn{2}{c|}{\textbf{Rack}} 
&   \multicolumn{2}{c|}{\textbf{Box}}    
& \multicolumn{2}{c|}{\textbf{Rack}}   
& \multicolumn{2}{c}{\textbf{Box}}  \\ 
\hline
\textbf{Training Method}       & \textbf{mIoU}      &     \textbf{mAP}           &  \textbf{mIoU}       & \textbf{mAP}         &   \textbf{mIoU}    &   \textbf{mAP}   &      \textbf{mIoU}   &   \textbf{mAP}   \\
\cline{1-9} 

Synthetic   &  $78.19$   &      $93.66$      & $77.30$     &  $91.43$    &   $85.47$  &   $97.69$   &   $81.01$   &  $94.34$     \\
Real  & $76.98$   &      $93.80$      & $80.07$     &  $92.20$    &  $88.42$  &  $97.79$   &   $81.20$   &  $93.04$         \\
Synthetic + Real  & $\textbf{95.05}$   &      $\textbf{97.46}$      & $\textbf{92.47}$     &  $\textbf{97.61}$      &   $\textbf{95.15}$  &  $\textbf{99.29}$   &   $\textbf{94.73}$  &  $\textbf{98.31}$     \\  
\end{tabular}
\end{adjustbox} 
\end{center}
\caption{\small{\textbf{Results on real-world data}: Here, we showcase the performance of \textit{Racklay-D-disc} on real-world data using three different training methods. For the first method (row 1), we train the model purely on synthetic data. For the second method (row 2), we train the model purely on real-world data. In the third method (row 3), we train the model on synthetic data and finetune with real-world data. The scores are scaled out of 100.}}
\label{table:real_world_results}

\end{table}


We test the transferability of our best network, \textit{Racklay-D-disc}, from synthetic images to real-world images using a custom dataset of 542 images \footnote{Download real-world dataset: \url{https://bit.ly/3BXvz4D}} which are manually annotated in the required format. We divide this data into 442 images for training and 100 images for testing. The data is also augmented using methods such as color jitter and  horizontal flip during training to increase the robustness of the model. We test the performance of the model on real world test images using 3 methods of training as described in Table \ref{table:real_world_results}:
\begin{enumerate}
    \item Using only synthetic data consisting of 8000 images (as explained in section \ref{sec:dataset}).
    \item Using only real world data of 442 images.
    \item Using a combination of the above two training sets: train the model on synthetic data and then fine-tune it using the real-world train images.
\end{enumerate}

The scores obtained using the first two methods are good but not at par with the results presented in Table \ref{table:quantitative:main}. Using a combination of synthetic and real world data for training, the model performs the best and the scores are at par with the scores in Table \ref{table:quantitative:main}. This not only shows that RackLay is able to transfer to real-world scenes effectively but also highlights the importance of the synthetic data generation pipeline. Due to the difficult nature of annotating the data, our synthetic data generation pipeline helps in generating large amounts of data which can be used to train the model. This model can then be finetuned with a comparatively smaller amount of real-world data to get good layouts. \textbf{We present the results for a few real-world images trained using method 3 in the supplementary material (\textit{real\_images.png}).}

\subsection{Applications} 
\subsubsection{\textbf{3D Free Space Estimation}}
We first obtain the 2D bounding boxes of all shelves and each box kept on it, from the top-view and front-view layouts. Considering a particular shelf, we then combine the corresponding 2D bounding boxes from $\widehat{\mathcal{T}}_{i}$ and  $\widehat{\mathcal{F}}_{i}$ to get respective 3D bounding boxes for each box stack on the shelf. Combining these representations for each shelf of a rack, we get a 3D volumetric reconstruction of the rack (refer Fig. \ref{fig:teaser_new}). We then calculate the total capacity of a shelf from the inter-shelf height obtained in the front-view layout and subtract the volumes of the reconstructed object stacks to obtain the free volume available on the particular shelf (refer Fig. \ref{fig:diversity}).
\subsubsection{\textbf{Counting Number of Boxes}}
After applying some morphological operations (either on top-view or front view), we get disjoint box layouts for each shelf. By counting the number of connected components, we compute the number of boxes kept on the particular shelf (refer Fig. \ref{fig:diversity}).

\section{Conclusion}
We propose \textit{RackLay}, which to the best of our knowledge is the first approach that provides multi-layered layout estimation, unlike previous approaches that assumed a single dominant plane. \textit{RackLay}'s versatility is showcased across a large diversity of warehouse scenes and is vastly superior to prior art baselines adapted for the same task.   We also release a flexible dataset generation pipeline \textit{WareSynth} that will aid future research for vision-based  tasks in warehouse scenarios.

\bibliographystyle{ACM-Reference-Format}
\bibliography{ICVGIP21-CameraReady-Template}

\appendix

\end{document}